\documentclass[10pt,twocolumn,letterpaper]{article}

\usepackage{cvpr}
\usepackage{times}
\usepackage{epsfig}
\usepackage{graphicx}
\usepackage{amsmath}
\usepackage{amssymb}

\usepackage{colortbl,booktabs}
\usepackage{tabularx}
\usepackage{multirow}  
\usepackage{multicol}
\usepackage{threeparttable} 

\usepackage[breaklinks=true,bookmarks=false]{hyperref}

\cvprfinalcopy 

\def\cvprPaperID{4115} 

\ifcvprfinal\fi
\begin{document}

\title{Intention Oriented Image Captions with Guiding Objects}

\author{Yue Zheng, Yali Li and Shengjin Wang\\
\\
Department of Electronic Engineering, Tsinghua University\\
{\tt\small zhengy17@mails.tsinghua.edu.cn, \{liyali13, wgsgj\}@tsinghua.edu.cn}
}

\maketitle
\thispagestyle{empty}

\begin{abstract}
Although existing image caption models can produce promising results using recurrent neural networks (RNNs), it is difficult to guarantee that an object we care about is contained in generated descriptions, for example in the case that the object is inconspicuous in the image. Problems become even harder when these objects did not appear in training stage. In this paper, we propose a novel approach for generating image captions with guiding objects (CGO). The CGO constrains the model to involve a human-concerned object when the object is in the image. CGO ensures that the object is in the generated description while maintaining fluency. Instead of generating the sequence from left to right, we start the description with a selected object and generate other parts of the sequence based on this object. To achieve this, we design a novel framework combining two LSTMs in opposite directions. We demonstrate the characteristics of our method on MSCOCO where we generate descriptions for each detected object in the images. With CGO, we can extend the ability of description to the objects being neglected in image caption labels and provide a set of more comprehensive and diverse descriptions for an image. CGO shows advantages when applied to the task of describing novel objects. We show experimental results on both MSCOCO and ImageNet datasets. Evaluations show that our method outperforms the state-of-the-art models in the task with average F1 75.8, leading to better descriptions in terms of both content accuracy and fluency. 
\end{abstract}


\begin{figure}[t]
\begin{center}
\includegraphics[scale=0.45]{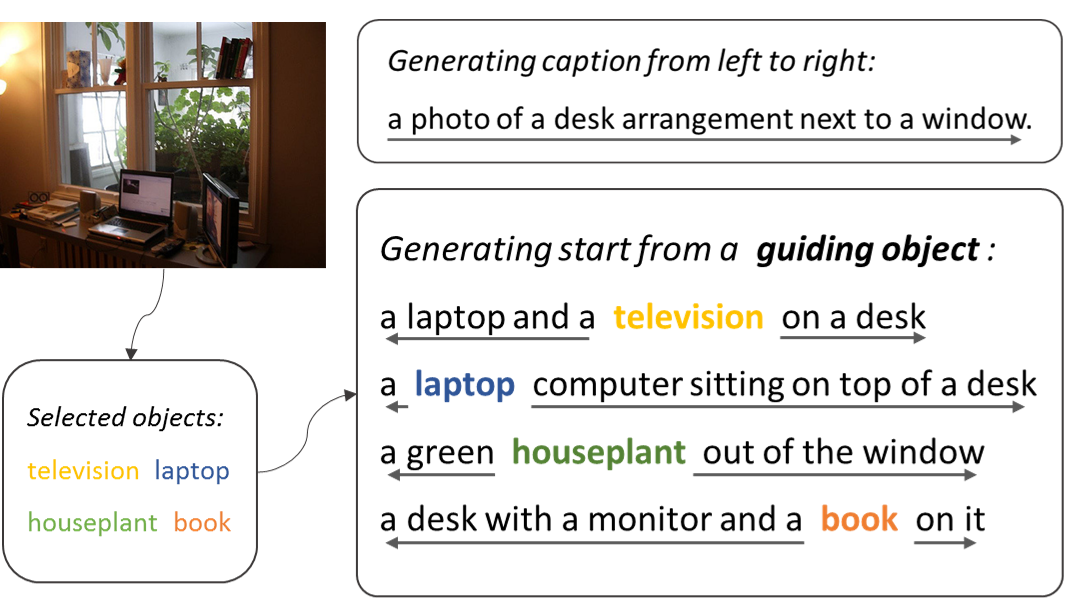}
\end{center}
   \caption{Existing models for image captioning generate the descriptions from left to right. Our CGO approach start generating with a selected object instead. CGO enables us to incorporate the selected objects into descriptions precisely, and generate a set of diverse and comprehensive descriptions for an image.}
\label{fig:dett}
\end{figure}

\section{Introduction}

Generating descriptions for images, namely image captioning, is a challenging task in computer vision. It can be used in many practical applications, such as robotic scene understanding and assistant systems for visually impaired users. In the past few years, deep neural networks are extensively used in image captioning \cite{mao2014deep,fang2015captions,vinyals2015show, karpathy2015deep,anderson2018bottom}, achieving fluent and accurate descriptions in commonly used datasets, \eg MSCOCO \cite{lin2014microsoft}. However, they are limited in the control of the generation process. For instance, a picture may contain many objects but a description sentence usually contains only one or a small number of objects, as shown in Fig. \ref{fig:dett}. Although we can accurately classify or detect objects in the image with existing methods \cite{he2016deep, simonyan2014very,ren2015faster}, we cannot force the language model to describe the object we care about. This can be important in practice because the model may be queried for a specific object. Including a novel object in the description is even harder, in the cases where the object has not been seen in the training data. Several recent works have studied the task of describing novel objects but it is still an open question. In this paper, we propose a novel approach that generates image captions with guiding objects (CGO). CGO can ensure that the user-selected guiding object is contained in a fluent description. The selected object can be any object detected from the image, even if it is unseen in the image caption training data.

The encoder-decoder structure is most widely used in recent image caption works and recurrent neural networks (RNNs) are often used as the language model to generate descriptions. In current approaches, the description is usually generated one by one as a word sequence from left to right. CGO is built on encoder-decoder structure, but instead of generating sequences from left to right, CGO generates sentences based on selected objects. We call them guiding objects as they guide the content of sequences in the generating process. A guiding object is the object that we want to include in the description. It may appear at any position in the sequence. We design a novel framework combining two LSTMs \cite{hochreiter1997long} to generate the left part and the right part of the sequence around the object. In this process, it is important that the content of the two sequences are coherent. In CGO, each LSTM encodes information of the other part of the sequence and then generates a sequence conditioned on the encoded sequence and visual features from the image. This helps the two sequences to connect with the guiding object fluently. It also enables us to generate multiple different descriptions for each selected object by providing different information sequences to the LSTMs.

Some earlier works on image caption tasks are template-based methods \cite{kulkarni2013babytalk,elliott2015describing}. These methods detect visual concepts from an image and fill them into templates directly. Although this enables us to control the presence of selected objects in the descriptions, the generated sentences are in limited forms and lack diversity. In CGO approach, the guiding object does not go through the encoding-decoding process and thus it acts like in template-based methods. At the same time, as the sequences on both sides are generated by the LSTMs, the sentences can be more fluent and diverse comparing with template methods. This makes CGO better at dealing with the novel object captioning task. 

In this paper, we first demonstrate the characteristics of our method on MSCOCO to generate descriptions for each detected object in the image. Usually only a small portion of objects in each image is mentioned in image caption labels. With CGO, however, we can extend the ability of description to the objects which are neglected and thus provide a set of more comprehensive and diverse descriptions for an image (as in Fig.\ref {fig:dett}). Then we apply CGO to the novel object captioning task and show its advantages when facing with unseen objects. We test our proposed approach on MSCOCO dataset and exhibit descriptions generated for ImageNet \cite{russakovsky2015imagenet} objects. Experiments show that our method outperforms the state-of-the-art models on multiple evaluation metrics, such as METEOR \cite{denkowski2014meteor}, CIDEr \cite{vedantam2015cider}, SPICE \cite{anderson2016spice} and novel object F1 scores. The generated descriptions are improved in terms of both content accuracy and fluency.

\section{Related Work}

\noindent {\bf Image captioning.}
In earlier image captioning studies, template-based models \cite{kulkarni2013babytalk,elliott2015describing} or retrieval-based models \cite{Devlin2015Language} were commonly used. The template-based models detect visual concepts from a given image and fill them into templates to form sentences. Thus the generations usually lack diversity. The retrieval-based models find the most consistent sentences from existing ones and cannot generate new descriptions. In recent works, the structure of encoder-decoder with deep neural networks is widely used \cite{vinyals2015show, karpathy2015deep}. In \cite{xu2015show, fang2015captions, lu2017knowing}, attention mechanism is used to make the language model pay attention to different areas of the image at each time step. In \cite{ranzato2015sequence, rennie2017self, liu2017improved, zhang2017actor}, reinforcement learning algorithms are applied to train the language models, enabling non-differentiable metrics to be used as training objectives.

\noindent {\bf Diverse descriptions.}
Controllability in generating process and diversity of descriptions are studied in recent years \cite{Bo2017Towards,jain2017creativity,shetty2017speaking,vijayakumar2016diverse,wang2017diverse,zhang2019more}. GAN-based methods \cite{Bo2017Towards,shetty2017speaking} and VAE-based methods \cite{jain2017creativity,wang2017diverse} are used to improve diversity and accuracy of descriptions. In \cite{mao2018show}, generated sentences can contain words of different topics. \cite{deshpande2018diverse} proposed a method to constrain the part-of-speech of words in generated sentences. Different from CGO, these methods do not precisely control the inclusion of objects in the descriptions.  \cite{chatterjee2018diverse,Bo2017Towards} studied generating descriptive paragraphs for images. \cite{johnson2016densecap} generates descriptions for each semantic informative region in images. These approaches require additional labels in dataset, \eg Visual Genome \cite{krishna2017visual}. CGO approach does not need additional labels. The descriptions are generated based on the whole image with CGO, so the objects may have richer relationships with each other. 

\noindent {\bf Describing novel objects.}
The novel object captioning task is first proposed by Hendricks \etal \cite{anne2016deep}. The proposed model DCC is required to describe objects unseen during training. In NOC \cite{venugopalan2017captioning}, a joint objective is used to train object classifiers and language models together. LSTM-C \cite{yao2017incorporating} applied copying mechanisms in NLP to incorporate novel words in generations. NBT and DNOC \cite{lu2018neural,wu2018decoupled} use language models to generate templates with slots or place-holders and then fill them with objects recognized from images. Unlike
\cite{anne2016deep, venugopalan2017captioning, yao2017incorporating, lu2018neural,wu2018decoupled}, novel objects are not predicted by the
language model in CGO, making it possible to contain novel
words in sentences precisely. CBS \cite{anderson2016guided} constrains the objects contained in generated sentences by adding constraints in beam search process. Different from CBS, novel words does not participate in the calculation of probability when decoding in CGO. Some works \cite{mou2016sequence} in NLP researches also use approaches of generating sentences with constrained words.


\begin{figure*}
\begin{center}
\includegraphics[scale=0.5]{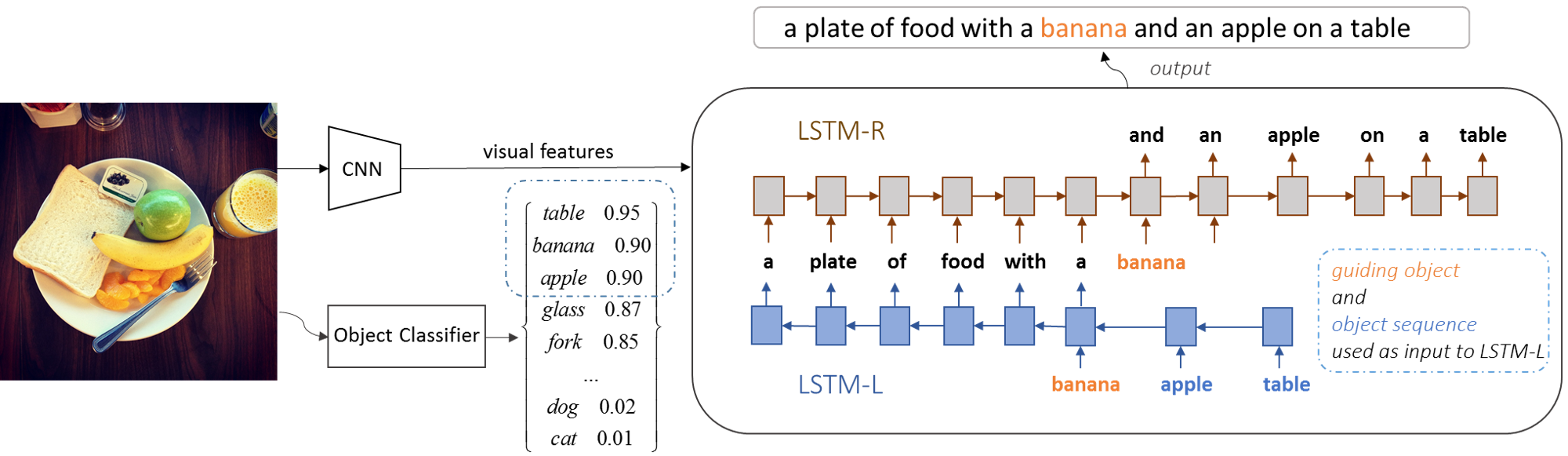}
\end{center}
   \caption{Our CGO approach. We select the guiding object and an object sequence according to the output of an object classifier. The sequence of objects is used as input to LSTM-L, providing with information about an assumed right-side sequence. LSTM-L generates the left-side sequence according to visual features and the input object sequence. The generated left-side sequence is then used as input to the LSTM-R to generate the right-side sequence. Then we connect the two partial sequences with the guiding object to get a complete description.}
\label{fig:short}
\end{figure*}

\section{Approach}

Given an image, CGO is able to incorporate the selected guiding object into generated sentences. In this process, two LSTMs are combined to generate the partial sequences on two sides of the guiding object. We use LSTM-L to denote the LSTM generating the left part of the sequence and LSTM-R to denote the other that generates the sequence on the right side. CGO can be applied flexibly to other existing RNN language models.
\subsection{Problem Formulation}
In commonly-used encoder-decoder models, convolutional neural networks (CNNs) \cite{simonyan2014very, he2016deep} are usually used as the encoder. Visual features representing information of the image from a CNN are then passed to a language model. RNNs such as the LSTM, are usually used as the language model in the decoding process. Given an image $I$, we aim to generate a sequence $ \mathbf{y} = (y_1, y_2, ..., y_T)$ for description, where $T$ denotes the length of the sequence and $y_i$ is a word in the model vocabulary. The size of the vocabulary is $V$. Denoting $\boldsymbol {\theta}$ the parameters in the encoder-decoder model, the objective of the learning process is to find the optimal $\boldsymbol {\theta}$ so that
\begin{equation}\boldsymbol{\theta^*} ={\rm argmax} \, \rm p(\mathbf {y^*}|I, \boldsymbol \theta)\end{equation}
where $\boldsymbol{\theta^*}$ denotes the optimized model parameters, $\mathbf {y^*}$ denotes the ground truth sequence. When the LSTM is used as the language model, at each time step $t$ it predicts the probability of the next word in the sequence according to image features $\mathbf{f_t}$, the input word $x_t$ at this time step and the hidden state $\mathbf{h_{t-1}}$ of the LSTM at time $t-1$. $x_t$ belongs to the model vocabulary.
\begin{equation} p(y_t|y_1,...,y_{t-1})=LSTM(\mathbf{f_t},x_t,\mathbf{h_{t-1}})\end{equation}
The image features $\mathbf{f_t}$ vary in different model settings. In some models, \eg NIC \cite{vinyals2015show}, image features $\mathbf{f}$ is only provided to the language model at time step $t = 0$. Models using attention mechanisms will use attended image features at each time step $t$ as
\begin{equation}\mathbf {a_t}=ATT(x_t,\mathbf{h_{t-1}})\end{equation}
\begin{equation}\mathbf{f_t}=\mathbf{f}\odot \mathbf{a_t}\end{equation}
where $\mathbf{a_t}$ denotes the attention weight maps at the time step $t$ and $\odot$ denotes element-wise multiplying. The form of the function $ATT$ to calculate attention weights varies with different attention mechanisms. 

If we hope the generated sequence contain a specific word, the desired sequence becomes $ \mathbf{y} = (y_1, ...,y_{k-1},y_k,y_{k+1},..., y_T)$, where $y_k$ is the specific word. At this time, the model output is conditioned on both image $I$ and the word $y_k$. Model parameters are trained to be
\begin{equation}\boldsymbol{\theta^*} ={\rm argmax} \, \rm p(\mathbf {y_{left}^*}|I, y_k, \boldsymbol \theta)\rm p(\mathbf {y_{right}^*}|I, y_k, \boldsymbol \theta)\end{equation}
where $\mathbf {y_{left}}= (y_1, ...,y_{k-1})$ and $\mathbf{y_{right}} = (y_{k+1}, ...,y_T)$. $\mathbf {y_{left}^*}$ and $\mathbf{y_{right}^*}$ are the ground truth partial sequences. The right-side partial sequence $\mathbf{y_{right}}$ could be of arbitrary length. We combine two LSTMs in opposite directions to complete the sequences on both sides of $y_k$.

\subsection{LSTM-L}
For the given image $I$ and the word $y_k$, we first use the LSTM-L to generate the left-side partial sequence. At each time step $t$, LSTM-L predicts the previous word conditioned on the image features $f_t$, the input word $x_t$, and the hidden state $h_{t+1}$.
\begin{equation}\ p(y_t|y_{t+1},...,y_{k})=LSTM_L(\mathbf{f_t},x_t,\mathbf{h_{t+1}})\end{equation}

\begin{figure}[t]
\begin{center}
\includegraphics[scale=0.35]{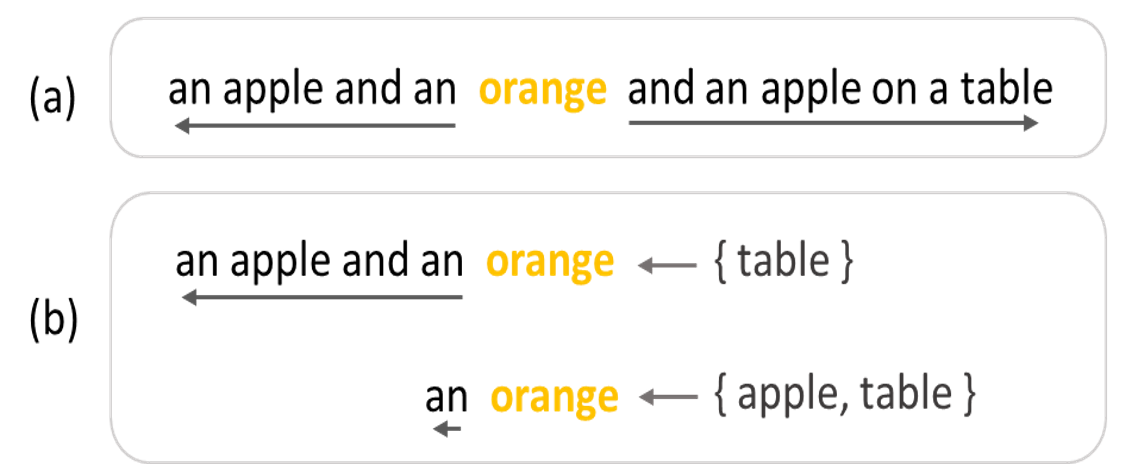}
\end{center}
   \caption{(a) Generated sequences on two sides of the guiding object (`orange') could be incoherent when they are generated independently. (b) Object label sequences are used as input to LSTM-L, providing with information about the right-side sequence. The LSTM-L generates different left-side sequences when the input sequences are different.}
\label{fig:methed2}
\end{figure}

However, there are problems in this process. An image often contains more than one objects. These objects can be arranged in descriptions in different orders. For instance, `there is an apple and a banana on the table' and `there is a banana and an apple on the table' are both correct descriptions. These two sentences could appear in the ground truth labels of an image at the same time. LSTM-L would have no idea about the right-side partial sequence when it is only provided with $y_k$ (Fig. \ref{fig:methed2}(a)). In experiments, we found that the model would tend to output a general and conservative results, such as `a banana' in such process. It is usually correct in grammar but lacking in variety. In contrast, various objects would occur in the left-side partial sequence in human generated descriptions.

The objects to be described are usually decided before we speak. Similarly in image captioning, we could get sufficient information about objects in the image before generating a description. Therefore, we first assume that a set of objects will appear in the description, and set the order in which these objects are arranged. Then we can get a sequence of object labels that is assumed to occur in the right-side sequence. We denote the object label sequence as $S=\{object_1,...,object_m\}$, where $m$ is the number of objects in $S$ and can be chosen arbitrarily. Objects in $S$ will not appear in the sequence generated by LSTM-L, but they will affect the contents in the sequence (Fig. \ref{fig:methed2}(b)). Sequence $S$ is used as input to LSTM-L and encoded before $y_k$. LSTM-L now generates the sequence according to the image $I$, the assumed sequence $S$ and $y_k$.
\begin{equation} p(y_t|y_{t+1},...,y_{k},S)=LSTM_L(\mathbf{f_t},x_t,\mathbf{h_{t+1}})\end{equation} 

Similar to normal generating processes, when the predicted word is the ending label $<$END$>$, the left-side sequence is completed and the sentence reaches the beginning.

At training time, we randomly select an object as $y_k$ from a ground truth caption label and then extract $S$ from the partial sentence on the right side of $y_k$. The left part of the sentence is provided to LSTM-L as the ground truth sequence. For a given image, we minimize the cross-entropy loss of the model.
\begin{equation}Loss=-\sum_{t=0}^{k-1} log\rm p (y_t^*|y_{t+1},...,y_k)\end{equation}
Note that the loss is only calculated for the generated left-side partial sequence, namely outputs at time steps earlier than $t=k$.

\subsection{LSTM-R}
After getting the left-side partial sequence from the LSTM-L, LSTM-R takes this sequence as input and complete the other part of the sentence. The model is now trained to be
\begin{equation}\boldsymbol{\theta_R^*} ={\rm argmax} \, \rm p(\mathbf {y_{right}^*}|I, y_{left}, y_k, \boldsymbol \theta )\end{equation}
In practice, we do not need to process the caption labels in the form as a right-side partial sequence. Instead, we can simply follow the process of training a normal LSTM that generates the sentences from left to right. The training loss for a given image and a selected $y_k$ is different between these two processes
\begin{equation}Loss_{normal}=-\sum_{t=0}^{T} log\rm p (y_t^*|y_{0},...,y_{t-1})\end{equation}
\begin{equation}Loss_{LSTM{-}R}=-\sum_{t=k+1}^{T} log\rm p (y_t^*|y_{0},...,y_{t-1})\end{equation}
where $Loss_{normal}$ and $Loss_{LSTM{-}R}$ denote loss functions in the two processes.
Note that $Loss_{normal}$ makes stricter restrictions than $Loss_{LSTM{-}R}$. The process of generating a complete sentence can be seen as a special case where the length of the input sequence is zero. On the other hand, the LSTM trained with complete sequences allows us to use the model more flexibly. When there is no object detected in the image (\eg a picture of the blue sky), or when no object is requested to be contained in the descriptions, we can use LSTM-R as a normal language model and start from the time step $t = 0$. In this case, the process is reduced to normal ones and generates sentences from left to right.

Our approach can be applied to all types of RNN language models for image captioning. In the inference process, various decoding methods can be used, including the greedy sampling and beam search methods.

\subsection{Novel Word Embedding}
In the encoding process, an input word $x_t$ is represented as a one-hot vector $\mathbf{x_t}$ and is then embedded with the learned parameter $W_x$. Embedding vector $W_x\mathbf{x_t}$ is used as input to the language model at time step $t$. A word that is unseen during training will not be generated by the language model when doing inference.

\begin{figure*}
\begin{center}
\includegraphics[scale=0.64]{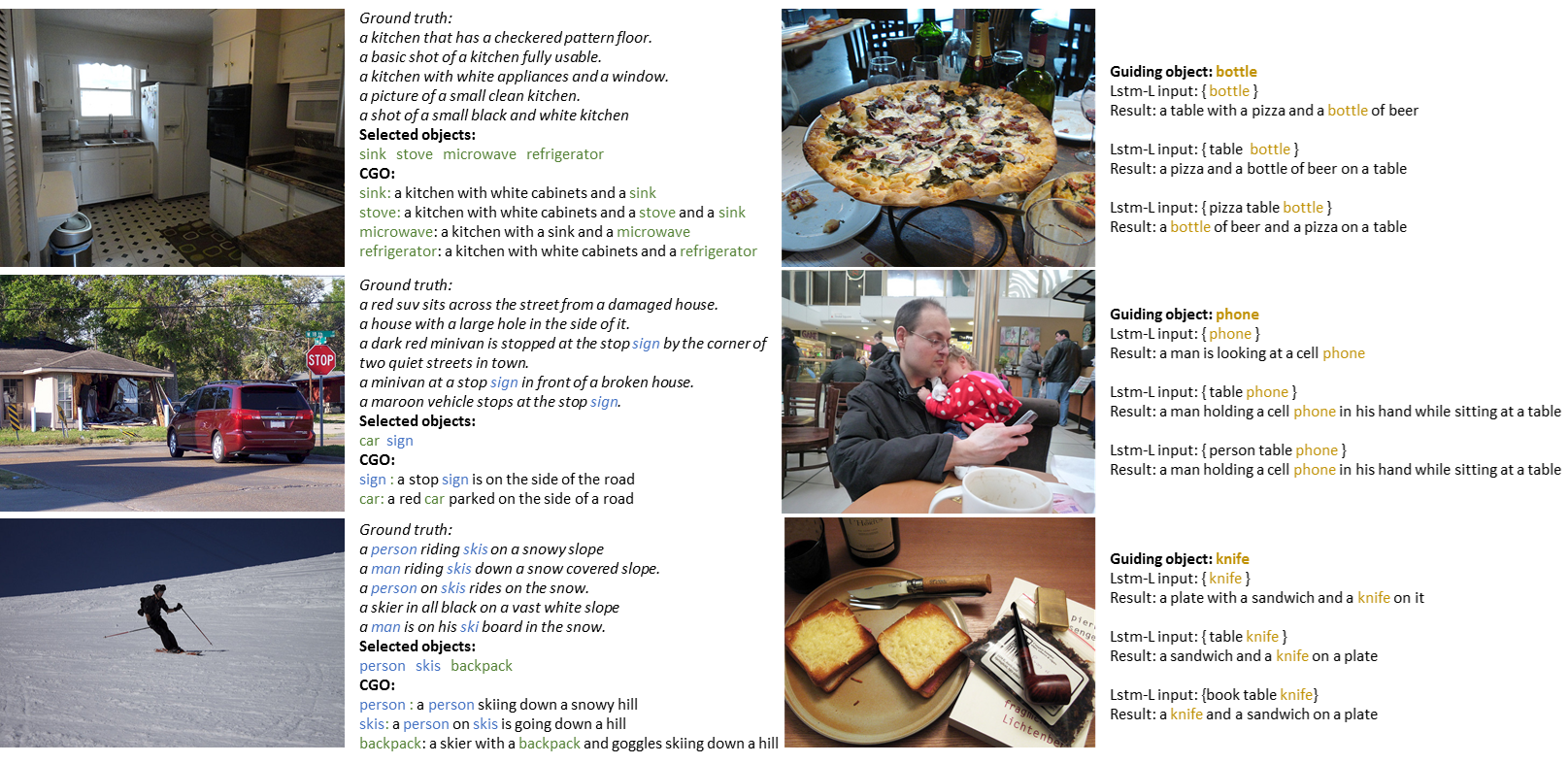}
\end{center}
   \caption{Examples of descriptions for selected objects are shown in the left column. The object at the beginning of each line in CGO results denotes the guiding word used for that description. Objects contained in the ground truth labels are in blue and the others are in green. Examples of diverse descriptions for a fixed guiding object is shown in the right column.}
\label{fig:det}
\end{figure*}

In CGO approach, when a novel object is selected as the guiding object, we can simply use the embedding vector from another seen object that is similar to this object. A similar object can be chosen according to WordNet \cite{miller1998wordnet} or distances between word embedding vectors from word2vec \cite{mikolov2013efficient} or GloVe \cite{pennington2014glove}. In normal left-to-right generating process, using embedding vectors from a similar object cannot force the language model to generate the novel word. With CGO however, since the novel word is incorporated in the generated sentence directly, without going through the encoding-decoding process, we do not need the language model to predict the novel word. Instead, the novel word is only used in the encoding process, and the embedding result from a similar object is sufficient in this process.

\subsection{Model Details}
\noindent {\bf Caption Model.}
In our experiments, we use the bottom-up and top-down attention model (Up-Down) \cite{anderson2018bottom} as our base model. The LSTM-L and the LSTM-R are both Up-Down models.  In our experiments, we use the pretrained model features from \cite{Anderson2017up-down}. It is extracted from a Faster R-CNN model built on ResNet-101 \cite{he2016deep} and is pretrained on MSCOCO and Visual Genome. 

\noindent {\bf Object Classifier.}
The objects in an given image can be recognized with existing object detection models or object classifiers. In our experiments, we follow previous works \cite{anne2016deep,anderson2016guided} using a  multi-label classifier to determine whether an object appears in the image. We classify the 80 object categories in the MSCOCO object detection dataset. We use the same feature in the classifier as in the language model.

\section{Experiments and Results}

In this section we show the ability of CGO to incorporate selected objects into descriptions. In subsection \ref{4.1} and \ref{4.2}, we show the characteristics of CGO by generating descriptions for each selected object in an image. In subsection \ref{4.2} we show the diversity of the generated descriptions. In subsection \ref{sec:ooc} and \ref{4.4} we apply CGO approach to the novel object captioning task. 

\noindent
{\bf Dataset.} Models are trained and evaluated on MSCOCO dataset which includes 123287 images. There are 80 object categories labeled for object detection and each image is labeled with 5 human generated descriptions for image captioning. We follow the previous work \cite{fang2015captions} to preprocess the caption labels that all labels are converted to lower case and tokenized. Words occur less than 5 times are filtered out and the others form a vocabulary of size 9487. We use the Karpathy's splits \cite{karpathy2015deep} in subsection \ref{4.1} and \ref{4.2} which is widely used in image caption studies. 113287 images are used in training set, 5000 images in validation set and 5000 in test set. In subsection \ref{sec:ooc} we use splits following \cite{anne2016deep}. Details are in subsection \ref{sec:ooc}. In subsection \ref{4.4} we test the model on the ILSVRC2012 validation set which contains 1000 classes and each image is labeled with its category.

\noindent
{\bf Training details.}
In our experiments, the object classifiers are optimized with stochastic gradient descent (SGD). The learning rate is set to 1e-4 and decays by 0.1 in every 10 epochs. The classifiers are trained for 20 epochs. The language models are optimized using Adam \cite{kingma2014adam}. The learning rate is set to 1e-4 and decays by 10 in every 20 epochs. The LSTM-L is trained for 80 epochs and LSTM-R is trained for 40 epochs.


\begin{table}[tp]  
\begin{center}  
  \begin{threeparttable}

    \begin{tabular}{lccc}  
    \toprule  
    Model&METEOR&Avg.Num&Avg.R\cr  
    \midrule  
    Base ( $b = 1$ )&26.6&1.50&0.55\cr 
    Base ( $b = 3$ )&27.3&1.68&0.59\cr
    Base ( $b = 5$ )&27.1&1.82&0.62\cr
    Base ( $b = 10$ )&26.7&1.98&0.66\cr
    \midrule 
    Base ( caption GT )&27.3&-&-\cr
    \midrule 
    CGO ( $k = 1$ )&24.4&1.62&0.50\cr  
    CGO ( $k = 3$ )&24.4&2.43&0.67\cr  
    CGO ( $k = 5$ )&24.2&2.77&0.73\cr
    CGO ( $k = 10$ )&24.2&2.92&0.75\cr
    \midrule
    Caption GT label&-&2.01&0.61\cr
    CGO ( caption GT )&28.0&-&-\cr  
    CGO ( det GT )&24.2&3.06&1.00\cr  
    \bottomrule  
    \end{tabular}  
    \end{threeparttable} 
\end{center}
\caption{`Base' denotes the base model used as baseline. $b$ indicates using the top $b$ beam search generations. For CGO we use the top $k$ objects predicted by the object classifier as guiding objects. `Caption GT label' is the statistical result of the ground truth labels. Base (caption GT) shows descriptions containing at least one object that occur in ground truth caption labels. CGO (caption GT) shows the score of descriptions whose guiding objects occur in ground truth caption labels. CGO (det GT) shows results when we generate descriptions for each object in the image (using object detection ground truth labels). Avg.Num denotes the average number of object categories in descriptions for an image. Avg.R denotes the average recall. }
 \label{tab:dettable}
\end{table} 

\begin{figure}[t]
\begin{center}
\includegraphics[scale=0.25]{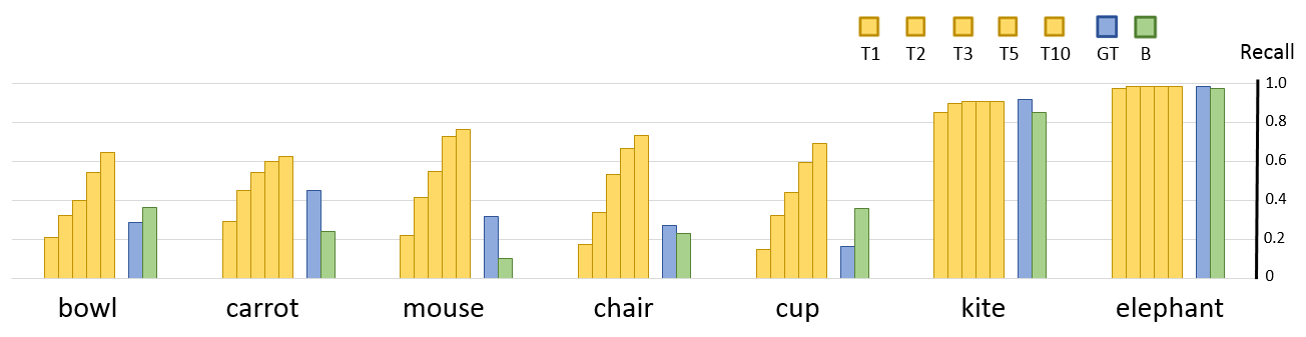}
\end{center}
   \caption{Examples of recalls for a single object category. T1$\sim$T10 denotes CGO results with different numbers of selected objects. GT denotes the statistical results of the ground truth caption labels. B denotes the results of the base model.}
 \label{fig:recall}
\end{figure}

\subsection{Describe Each Selected Object}
\label{4.1}

To demonstrate the characteristics of our method, we generate one description for each object selected in images to get a set of sentences describing different objects in each image. We choose $k$ objects with the highest probability as the guiding objects according to the outputs of the object classifier. We test the models with $k = 1, 3, 5, 10$ respectively. We count the average number of different object categories involved in the set of descriptions for each image. We also count the recall of each object category. That is, whether the object appearing in an image is mentioned in the set of descriptions. It should be noted that whether an object occurs in the image is decided according to the object detection labels, since caption labels often contain only a small portion of the objects appearing in the image.

\begin{table}[tp]  
\begin{center}  
  \begin{threeparttable}  

    \begin{tabular}{lcccc}  
    \toprule  
    &\multicolumn{2}{c}{2 objects selected}&\multicolumn{2}{c}{3 objects selected}\cr  
    \cmidrule(lr){2-3} \cmidrule(lr){4-5}  
    Object label&Uniq.&M&Uniq.&M\cr  
    \midrule  
    Caption label&1.47&26.1&2.08&25.7\cr 
    Detection label&1.48&24.7&2.62&23.5\cr  
    \bottomrule  
    \end{tabular}  
    \end{threeparttable} 
\end{center}
\caption{Results when we select guiding objects from caption labels and object detection labels. M denotes the METEOR score. Uniq. denotes the average number of unique descriptions for each fixed guiding object. }
  \label{tab:det table1} 
\end{table}

Both the base model and CGO are trained and tested on MSCOCO Karpathy's splits \cite{karpathy2015deep}. Examples are shown in the left column of Fig. \ref{fig:det}. 
In Table \ref {tab:dettable}, we show results generated with the base model using beam search (beam size = $b$) as baseline, and results generated with CGO. The average recall is the macro average of recalls for all 80 object categories. The average object category number and recall of baseline model are 1.98 and 0.66 with $b=10$. With CGO, the average number and recall are improved to 2.92 and 0.75 ($k=10$). We also count the average number and recall for the ground truth caption labels. When $k=5$ (there are 5 caption labels for each image), the object recall of CGO is 0.73, higher than that of the caption labels (0.61), indicating that CGO can describe the objects which are neglected in caption labels. Note that although the base model can describe more object categories with larger beam size, it cannot control which objects are described in the process.

The METEOR scores of CGO is around 24.2, lower than that of the base model (26.7). The evaluation method only compare the results with ground truth caption labels. Even if objects appearing in the image are described correctly, the scores would be low if the objects do not appear in the ground truth captions. Though the fluency of the generations cannot be evaluated precisely using this score, this provides us with a rough reference. We also evaluate the descriptions whose guiding objects appear in the ground truth labels and the METEOR score is 28.0. This suggests that the generated sentences are fluent when the guiding objects are in-domain for the caption labels.

\renewcommand{\arraystretch}{0.95}
\begin{table*}[tp]  
\begin{center}  
  \begin{threeparttable}  
   
    \begin{tabular}{lccccccccc}  
    \toprule    
    Model&bottle&bus&couch&microwave&pizza&racket&suitcase&zebra&Avg. F1\cr  
    \cmidrule(lr){1-9} \cmidrule(lr){10-10}  
    DCC \cite{anne2016deep}&4.6&29.8&45.9&28.1&64.6&52.2&13.2&79.9&39.8\cr 
    NOC \cite{venugopalan2017captioning}&17.8&68.8&25.6&24.7&69.3&68.1&39.9&89.0&49.1\cr  
    LSTM-C \cite{yao2017incorporating}&29.7&74.4&38.8&27.8&68.2&70.3&44.8&91.4&55.7\cr
    CBS+T4 \cite{anderson2016guided}&16.3&67.8&48.2&29.7&77.2&57.1&49.9&85.7&54.0\cr
    NBT + G \cite{lu2018neural}&14.0&74.8&42.8&63.7&74.4&19.0&44.5&92.0&53.2\cr 
    DNOC \cite{wu2018decoupled}&33.0&76.9&54.0&46.6&75.8&33.0&59.5&84.6&57.9\cr
    CGO (ours)&{\bf 45.0}&{\bf 79.0}&{\bf 69.2}&{\bf 64.6}&{\bf 87.3}&{\bf 89.7}&{\bf 75.8}&{\bf 95.0}&{\bf 75.8}\cr   
    \bottomrule  
    \end{tabular}  
    \end{threeparttable} 
\end{center}
\caption{F1 scores of the novel objects on the test split.}
  \label{tab:ooc table}
\end{table*}  

\begin{table*}[tp]  
\begin{center}  
  \begin{threeparttable}  
    
    \begin{tabular}{lccccccc}  
    \toprule  
    &\multicolumn{4}{c}{Out-of-Domain Scores}&\multicolumn{3}{c}{In-Domain Scores}\cr  
    \cmidrule(lr){2-5} \cmidrule(lr){6-8}  
    Model&SPICE&METEOR&CIDEr&Avg. F1&SPICE&METEOR&CIDEr\cr  
    \midrule  
    DCC \cite{anne2016deep}&13.4&21.0&59.1&39.8&15.9&23.0&77.2\cr 
    NOC \cite{venugopalan2017captioning}&-&21.4&-&49.1&-&-&-\cr  
    LSTM-C \cite{yao2017incorporating}&-&23.0&-&55.7&-&-&-\cr  
    CBS + T4 \cite{anderson2016guided}&15.9&23.3&77.9&54.0&18.0&24.5&86.3\cr
    NBT + G \cite{lu2018neural}&16.6&23.9&84.0&53.2&18.4&25.3&94.0\cr 
    \midrule 
    CGO ($p_o$ = 0.5)&17.7&23.9&89.1&{\bf 75.8}&18.0&25.1&94.7\cr  
    CGO ($p_o$ = 0.7)&17.7&23.9&88.2&{\bf 75.8}&18.4&25.3&95.8\cr  
    CGO ($p_o$ = 0.9)&{\bf 18.1}&{\bf 24.2}&{\bf 90.0}&{\bf 75.8}&{\bf 19.6}&{\bf 26.3}&{\bf 103.3}\cr  
    \bottomrule  
    \end{tabular}  
    \end{threeparttable} 
\end{center}
\caption{Descriptions generated for in-domain and out-of-domain images are evaluated using image caption metrics. $p_o$ is the threshold for selecting guiding objects which are in-domain. When the probability of occurrence of an object predicted by the object classifier exceeds $p_o$, it is used as the guiding object. We choose the object with the highest probability when more than one object meet the requirement.}
  \label{tab:ooc table1}
\end{table*} 
\renewcommand{\arraystretch}{1}

Figure \ref{fig:recall} shows recalls of 7 objects as examples. Comparing with the base model and ground truth caption labels, recall of inconspicuous objects such as ``cup'' (from 0.15 to 0.69) and ``bowl'' (from 0.21 to 0.65) can be improved significantly with CGO.


\subsection{Diverse Descriptions for Each Object}
\label{4.2}

In this part we show CGO's ability to generate diverse descriptions with a fixed guiding object. We randomly select an object as the guiding object from an image and choose $n = 1$ or $2$ other objects to form the LSTM-L input sequence. With $n=1$, the input sequence can be $<$Guiding object$>$ or $<$Guiding object, Object1$>$, `Object1' denotes the chosen object for the LSTM-L input sequence. With $n=2$ we test with 3 different input sequences with length 1, 2 and 3.

Results are shown in Table \ref {tab:det table1}. When we use objects chosen from object detection labels, the average number of unique descriptions is 1.48 with 2 different inputs, and 2.62 with 3 different inputs. This shows that CGO can generate different descriptions even with a fixed guiding object. Examples are shown in the right column in Fig. \ref {fig:det}. 

\subsection{Novel Object Captioning}
\label{sec:ooc}

In this part, we demonstrate the effectiveness of CGO when applied to the novel object captioning task. Following \cite{anne2016deep}, 8 object categories, `bus, bottle, pizza, microwave, couch, suitcase, racket, zebra' are selected as novel objects. At training time, images are excluded from the MSCOCO training set if their caption labels contain the novel objects. Half of the MSCOCO validation set is used as validation set and the other half as the test set. F1 score is used to evaluate the accuracy of containing the novel objects. For each novel object category, if the generated description and the ground truth label contain the object at the same time, it is regarded as true positive. The average F1 score is the macro average across the 8 categories. Image caption evaluating metrics are used to evaluate the quality of the generated sentences, including SPICE \cite{anderson2016spice}, METEOR \cite{denkowski2014meteor} and CIDEr \cite{vedantam2015cider}. Descriptions for out-of-domain images (containing a novel object) and in-domain images are evaluated respectively.

\begin{figure}[t]
\begin{center}
\includegraphics[scale=0.45]{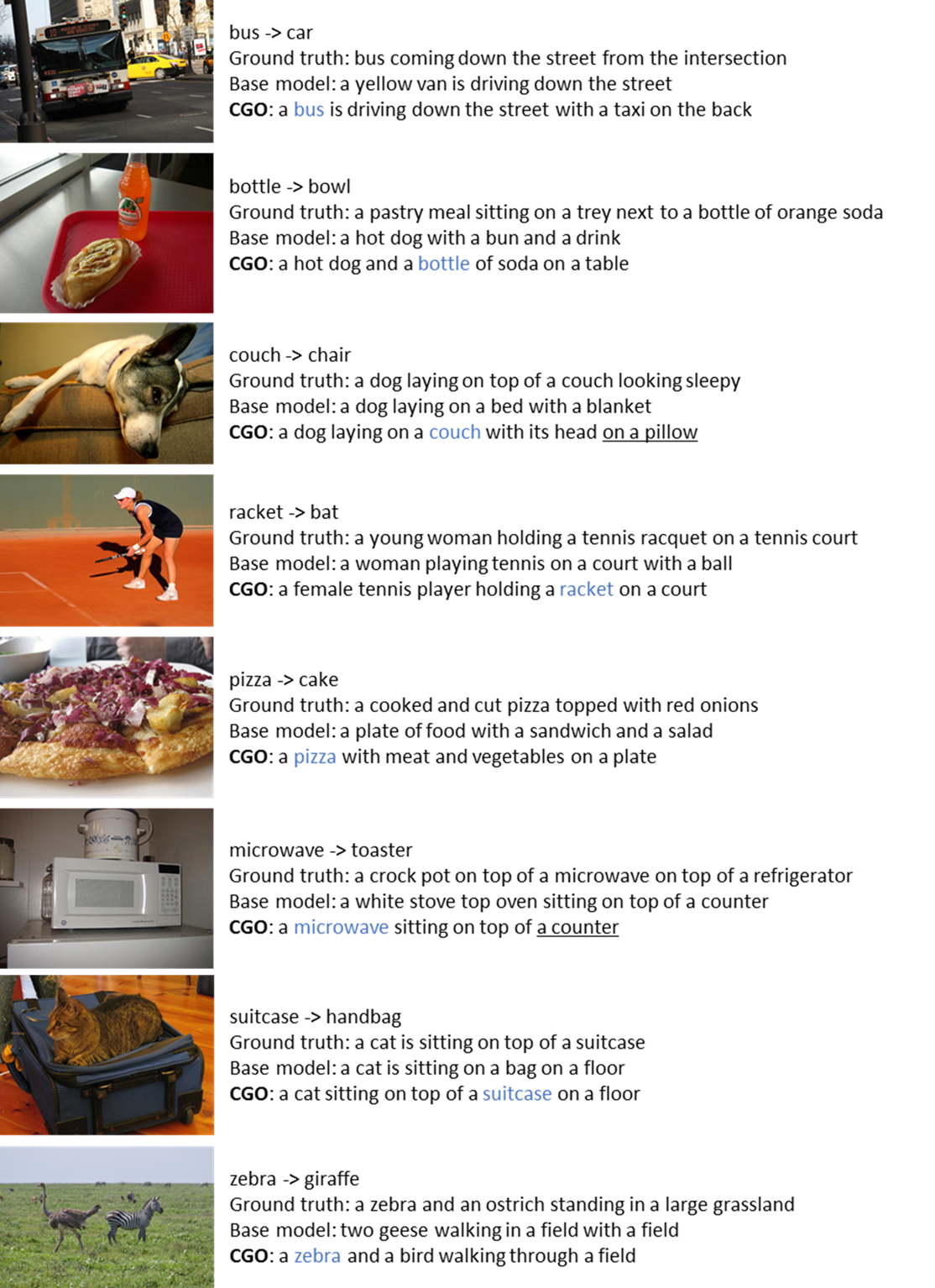}
\end{center}
   \caption{Examples of descriptions generated for out-of-domain objects (in blue). $\rm O_1\to \rm O_2$ indicates that we use $\rm O_1$ as the guiding object which is encoded using the word embedding vector of an in-domain object $\rm O_2$. Errors are underlined. }
\label{fig:ooc}
\end{figure}

Similar with prior work \cite{anne2016deep, anderson2016guided}, labels for the object classifier is obtained from caption labels. The full training set is used when training the classifier, including the images which contain novel objects. A novel object is used as the guiding object if it appears in the image. We determine whether a novel object appears in an image according to the results from the object classifier. The probability thresholds of using an object as the guiding object are chosen to maximize the F1 score on the validation set. For novel words, we simply replace their word embedding vectors with other in-domain objects under the same super categories, \eg `bottle' $\to$ `cup'. The results are shown in Table \ref {tab:ooc table} and Table \ref {tab:ooc table1}. Examples of descriptions are shown in Fig. \ref {fig:ooc}.

\begin{figure}[t]
\begin{center}
\includegraphics[scale=0.58]{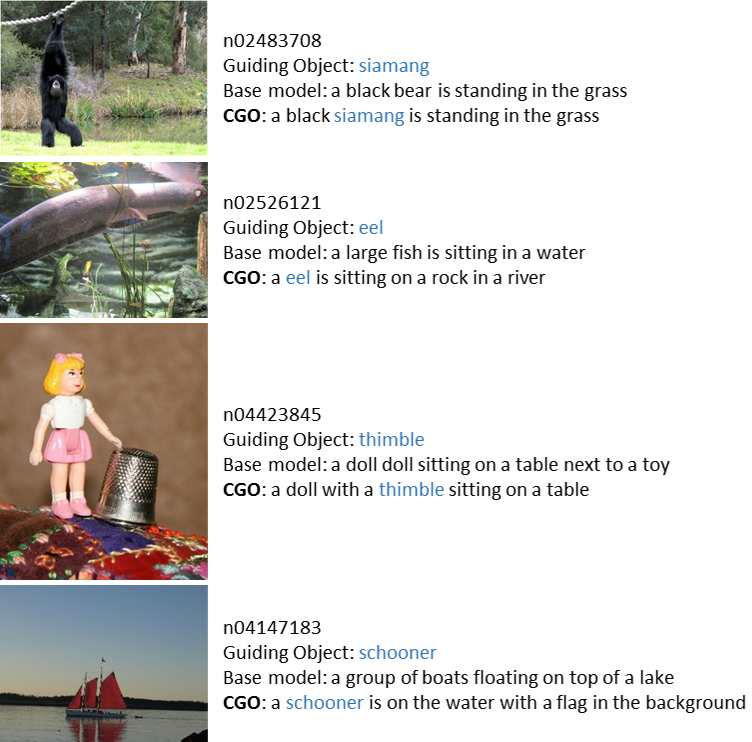}
\end{center}
   \caption{Examples of generated captions for ImageNet objects (in blue). With CGO, an object could be involved in descriptions even if it did not appear in the model vocabulary.}
\label{fig:ooc1}
\end{figure}

Comparing to existing models, CGO significantly improves the F1 score of novel objects, with the average F1 score 75.8. In fact, the F1 score of the output directly depends on the accuracy of the classification results, just like the template-based models. Note that using different RNN models or different CNN features in the language model does not affect the F1 result. On the other hand, CGO takes advantage of the LSTM language model and generates fluent descriptions. METEOR scores are improved to 24.2 for out-of-domain images and 26.3 for in-domain images. We test the CGO with different probability thresholds $p_o$ for in-do-main objects. An in-domain object is used as the guiding object when probability predicated by the object classifier exceed the threshold. The generating process is reduced to the usual left-to-right generation process when the classifier is not certain about the objects contained in the image.

As the object classifier is independent with the language model, using more advanced models such as object detection models might further improve the F1 scores. In CGO approach, we only guarantee one selected object to be mentioned, but this does not affect its practicability. In many scenarios, novel words do not appear intensively and we are allowed to use more than one description for an image in practice. In addition, CGO can be used in conjunction with other methods such as CBS \cite{anderson2016guided}, to contain more objects in the outputs while ensuring that the guiding object is contained in the descriptions.

\subsection{Descriptions for ImageNet Objects}
\label{4.4}

Similar to previous works \cite{yao2017incorporating, venugopalan2017captioning}, we show qualitative results of our method describing the ImageNet \cite{russakovsky2015imagenet} objects. Objects which do not appear in the vocabulary mined from the MSCOCO caption labels are novel to the models trained on the MSCOCO. We use the model trained on Karpathy's training split to generate descriptions. Examples are shown in Fig. \ref {fig:ooc1} and more examples of results can be found in Appendix. Similar to the process in subsection \ref {sec:ooc}, word embedding vectors of novel objects are replaced with embedding vectors of the seen objects. \eg `schooner'$\to$`boat'. 

\section{Conclusion}

We present a novel image captioning approach where the sentence generating process starts from a selected guiding object. Our CGO allows us to include a specific object in generated sentences and describe images in a comprehensive and diverse manner. 

\section*{Acknowledgements}

This work was supported by the National Natural Science Foundation of China under Grant Nos. 61701277, 61771288 and the state key development program in 13th Five-Year under Grant No. 2016YFB0801301.

{\small
\bibliographystyle{ieee_fullname}
\bibliography{suo0}
}

\end{document}